\documentclass{article}

\usepackage{microtype}
\usepackage{graphicx}
\usepackage{framed}
\usepackage{pifont}
\usepackage{subfigure}
\usepackage{algorithm}
\usepackage{algorithmic}
\usepackage{booktabs} 

\usepackage{hyperref}

\usepackage[accepted]{icml2024}

\usepackage{amsmath}
\usepackage{amssymb}
\usepackage{asymptote}
\usepackage{pifont}
\usepackage{mathtools}
\usepackage{amsthm}
\usepackage{xcolor} 
\usepackage{colortbl} 
\usepackage[capitalize,noabbrev]{cleveref}

\theoremstyle{plain}

\theoremstyle{definition}

\theoremstyle{remark}

\usepackage[textsize=tiny]{todonotes}

\icmltitlerunning{  }
\hypersetup{
    colorlinks=true, 
    linkcolor=red,   
    citecolor=blue   
}
\begin{document}

\twocolumn[{
\icmltitle{MACM: Utilizing a Multi-Agent System for Condition Mining in Solving Complex Mathematical Problems}



\icmlsetsymbol{equal}{*}

\begin{icmlauthorlist}
\icmlauthor{Bin Lei}{sch}
\icmlauthor{Yi Zhang}{sch}
\icmlauthor{Shan Zuo}{sch}
\icmlauthor{Ali Payani}{comp}
\icmlauthor{Caiwen Ding}{sch}
\end{icmlauthorlist}

\icmlaffiliation{sch}{Department of Computer Science and Engineering, University of Connecticut, Storr, USA}
\icmlaffiliation{comp}{Cisco, Location, Country}

\icmlcorrespondingauthor{Firstname1 Lastname1}{first1.last1@xxx.edu}
\icmlcorrespondingauthor{Firstname2 Lastname2}{first2.last2@www.uk}

\icmlkeywords{Machine Learning, ICML}

\vskip 0.3in

}
]







\begin{abstract}
Recent advancements in large language models, such as GPT-4, have demonstrated remarkable capabilities in processing standard queries. Despite these advancements, their performance substantially declines in \textbf{advanced mathematical problems requiring complex, multi-step logical reasoning}. To enhance their inferential capabilities, current research has delved into \textit{prompting engineering}, exemplified by methodologies such as the Tree of Thought and Graph of Thought.
Nonetheless, these existing approaches encounter two significant limitations. Firstly, their effectiveness in tackling complex mathematical problems is somewhat constrained. Secondly, the necessity to design distinct prompts for individual problems hampers their generalizability.
In response to these limitations, this paper introduces the \textit{Multi-Agent System for conditional Mining} (\textbf{MACM}) prompting method. It not only resolves intricate mathematical problems but also demonstrates strong generalization capabilities across various mathematical contexts.
With the assistance of MACM, the accuracy of GPT-4 Turbo on the most challenging level five mathematical problems in the MATH dataset increase from $\mathbf{54.68\%}  \text{ to } \mathbf{76.73\%}$. The code is available in \url{https://github.com/bin123apple/MACM}.
\end{abstract}

\section{Introduction}
Large language models (LLM), such as GPT-4~\cite{OpenAIGPT4}, have demonstrated exceptional capabilities in handling a wide range of problems. However, despite their proficiency in understanding language and generating text, these models still exhibit certain limitations when dealing with problems involving complex logical deduction~\cite{nori2023capabilities}, especially for mathematical problems that involve abstract concepts and multi-step logical derivations~\cite{abramski2023cognitive}. This shortfall primarily results in these models' inability to consistently reach the level of accuracy and reliability expected in fields that require precise mathematical reasoning, such as advanced academic research, engineering problem-solving, and theoretical physics.  

A contemporary and efficacious method for tackling this issue is the prompting engineering~\cite{white2023prompt}. It enhances accuracy in complex problem-solving without necessitating further training of the model. By strategically crafting prompts, prompting engineering optimizes the utilization of large language models, guiding their processing pathways more efficiently and effectively~\cite{liu2023pre}. 
\begin{figure}[h]  
\centering 
\includegraphics[width=0.46\textwidth]{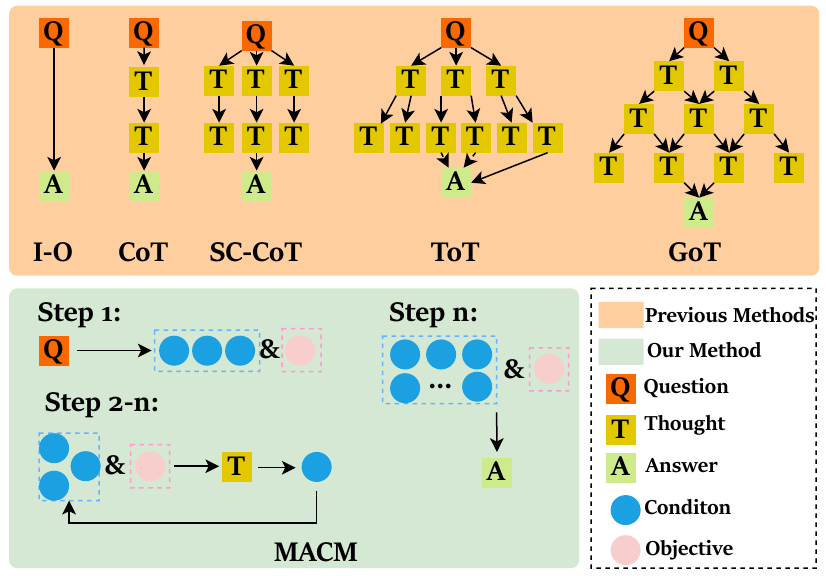} 
\caption{The Comparison Between the  Current Mainstream Prompting Methods and MACM. MACM extracts conditions and the objective from each math problem, iteratively adds new insights to the known conditions, and repeats this until enough information is gathered to reach a solution.}  
\label{fig:Basic_flow}
\end{figure}

Previous prompting methods mainly include the Chain of Thought (CoT)~\cite{wei2022chain}, Self-consistency Chain of Thought (SC-CoT)~\cite{wang2022self}, Tree of Thought (ToT)~\cite{yao2023tree}, and Graph of Thought (GoT)~\cite{besta2023graph,lei2023boosting}. However, the CoT and SC-CoT methods exhibit low capabilities in complex logical reasoning. For instance, even when applied to relatively simple challenges like the 24-point game, CoT and SC-CoT only achieve accuracy rates of 4.0\% and 9.0\% by utilizing the GPT-4~\cite{yao2023tree}, respectively. Although the ToT and
GoT methods have enhanced the ability of LLMs to process complex logical problems, they lack sufficient generalizability. These methods require setting specific prompts for each particular problem, as detailed in the Appendix~\ref{Appendix:generalize}.

To address two key issues:
\vspace{-5pt}
\begin{enumerate}
    \item The insufficient reasoning capability of LLMs for complex mathematical problems;
 space between items
    \item The inadequate generalizability of current prompting methods;
\end{enumerate}
We propose the \textit{Multi-Agent System for Condition Mining (\textbf{MACM})} prompting method. MACM  has moved beyond being restricted by the specific contents of a problem. Instead, it first abstracts the conditions and the objective of the problem. Subsequently, through a Multi-Agent interactive system, it progressively mines new conditions conducive to achieving the objective, thereby ultimately fulfilling the goal of problem-solving.

The comparison between MACM and the basic approaches of current mainstream prompting methods in problem-solving is illustrated in Figure~\ref{fig:Basic_flow}. Our method abandons the dependency on hierarchical relationships prevalent in previous methods and introduces the concepts of Conditions and the Objective. It works by continually expanding the existing known conditions to yield the final Answer. 
The introduction of the concepts of Conditions and Objectives in MACM eliminates the need to manually set specific prompts for each particular problem. Continuously updating the known Conditions is akin to compressing all information from various Thoughts into the existing Conditions. This approach enables our method to capture more connections between Thoughts than previous prompting methods, which rely on traversing and filtering through a hierarchical structure. For a detailed description of the method and prompts, refer to Section~\ref{Sec:Method}.

Through comparisons in three experiments on the MATH dataset, the 24-point game, and number sequence sorting, we have verified the generalizability of MACM and its superior error correction capability compared to original prompting methods. With our method, the accuracy of the GPT-4 turbo model on the MATH dataset increased by $\mathbf{15.14\%}$. Compared to SC-CoT, its accuracy improved by $\mathbf{7.8\%}$. In the 24-point game, using the same GPT-4 model, MACM achieved an accuracy rate $\mathbf{17\%}$ higher than ToT.
\section{Related Work}

In this section, we provide a summary of several major current prompting methods.

\textit{I-O} Prompting:
Input-Output (I-O) prompting stands as the cornerstone of interaction with large language models, epitomizing the most widespread method for eliciting responses from these AI systems. In this methodology, a user specifies the problem conditions directly to the model, which then embarks on generating answers through a token-level, sequential decision-making process that unfolds from left to right \cite{zhang2022automatic}. As the foundational approach embedded within the operational mechanics of models like GPT-4, I-O prompting leverages the inherent structure of language models to interpret the input and produce contextually relevant outputs.

\textit{CoT} Prompting \cite{wei2022chain}:
Chain of Thought (CoT) prompting is designed to refine the model's output into more structured and logically coherent text, fostering a sense of logical progression in its responses. This methodology operates on the principle that by methodically constructing and elaborating upon chains of reasoning, the model is better equipped to produce outputs that are not only coherent but also deeply rooted in logical deduction. CoT prompting thereby encourages the model to engage in a more deliberate and sequential reasoning process when addressing queries. By explicitly guiding the model to connect each step in its reasoning, it ensures that the final output maintains a logical flow and coherence. This approach significantly enhances the quality of the model's responses, making it particularly effective for complex problem-solving tasks that demand a high degree of logical rigor and analytical depth.

\textit{SC-CoT} Prompting \cite{wang2022self}:
Self-Consistency Chain of Thought (SC-CoT) prompting represents an advanced iteration of the Chain-of-Thought (CoT) technique, enhancing the model's capacity for producing text that is not only logically structured but also self-consistent. This methodology underscores the importance of generating content that maintains internal consistency and semantic interconnectedness across all parts of the text. Achieving such a high level of consistency and logical coherence is facilitated through a sophisticated process wherein large models are employed to evaluate (vote or score) their own outputs, which is then followed by a meticulous selection of the most coherent response. By prioritizing self-consistency within the model's outputs, SC-CoT prompting significantly mitigates the occurrence of logical fallacies and inconsistencies, thereby elevating the overall quality and reliability of the text generated. This methodological advancement empowers models to handle complex textual generation tasks with greater precision and fidelity to logical coherence, marking a substantial leap forward in the application of AI in areas requiring nuanced and consistent reasoning.

\textit{ToT} Prompting \cite{yao2023tree}:
Tree of Thought (ToT) prompting introduces a novel approach by implementing structured prompts that guide the model in crafting text that is both hierarchical and systematically organized. This method capitalizes on a tree-like conceptual framework to elucidate the connections among various ideas, employing this structured representation as the basis for input prompts. Through this, the model is endowed with the ability to methodically organize and deliberate on the text in alignment with the tree's architecture, fostering the generation of responses that are not only more precise but also inherently structured. ToT further enhances its methodology with a voting or scoring mechanism, streamlining the process to refine the outcomes. This not only aids in minimizing errors but also contributes to reducing the computational demands. The overarching goal of Tree-of-Thought prompting lies in its capacity to enrich semantic representation and bolster logical structuring, significantly amplifying the model's deductive reasoning and quality of output, thereby setting a new precedent in the utilization of structured thinking paradigms within AI-generated content.

\textit{GoT} Prompting \cite{besta2023graph,lei2023boosting}:
GoT builds upon the foundation of ToT by enhancing the connections between thoughts, making it possible for thoughts located on different branches of the original tree structure to interconnect. Additionally, GoT decomposes complex tasks into simpler sub-tasks, solves these sub-tasks independently, and then incrementally merges these outcomes into the final result, thereby reducing the cost.
\section{Method}
\label{Sec:Method}
\subsection{MACM  Overall Structure}
\begin{figure*}[ht]  
\centering 
\includegraphics[width=0.99\textwidth]{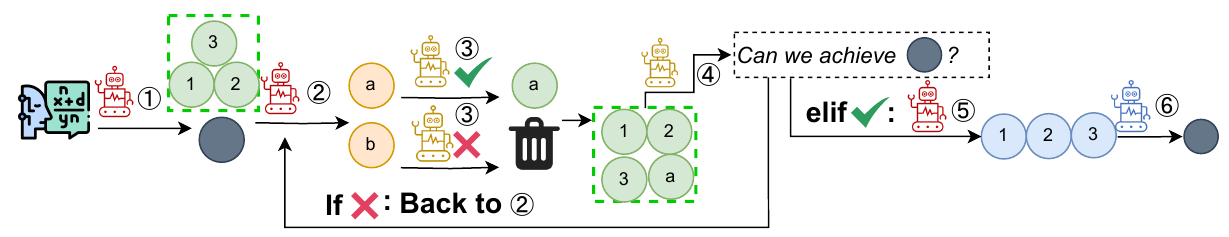} 
\caption{The overall structure of MACM. \includegraphics[height=1em]{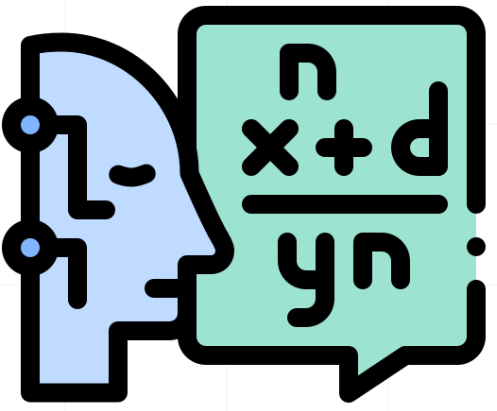}: Original Math problem; \includegraphics[height=1em]{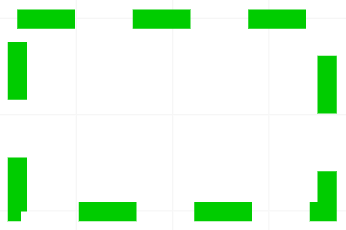}: Condition list; \includegraphics[height=1em]{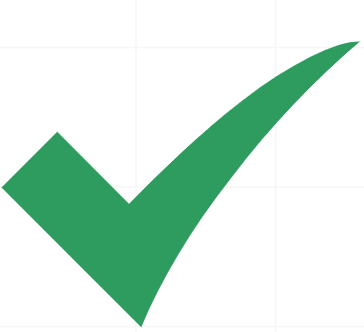}: True; \includegraphics[height=1em]{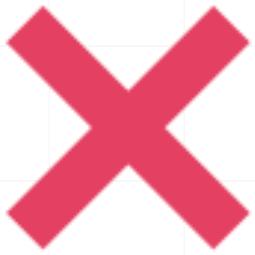}: False; \includegraphics[height=1em]{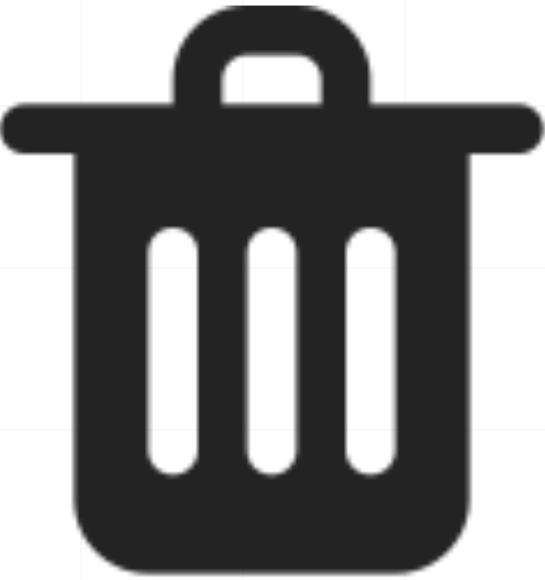}: Discard; \includegraphics[height=1em]{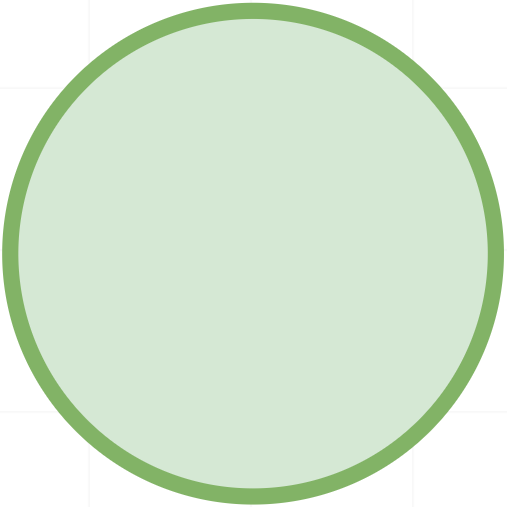}: Known Conditions; \includegraphics[height=1em]{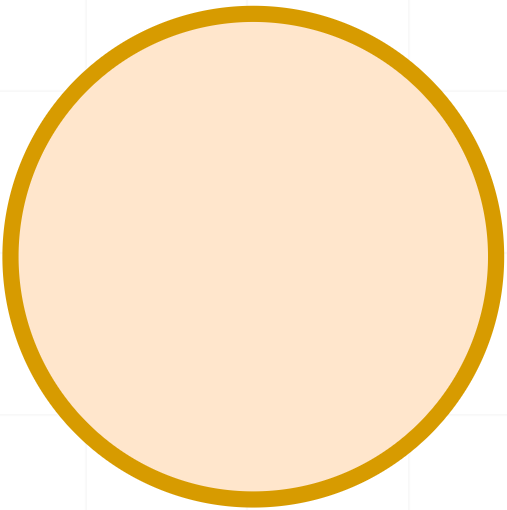}: New Conditions; \includegraphics[height=1em]{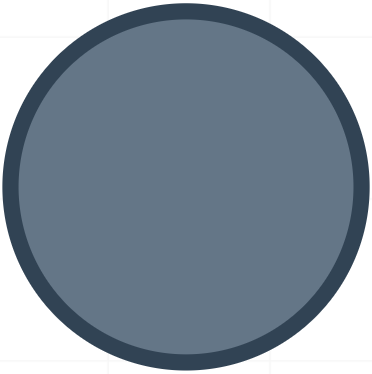}: Objective; \includegraphics[height=1em]{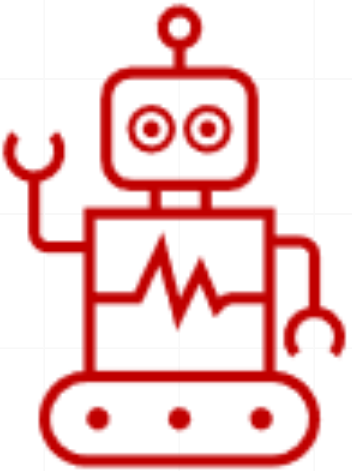}: Thinker; \includegraphics[height=1em]{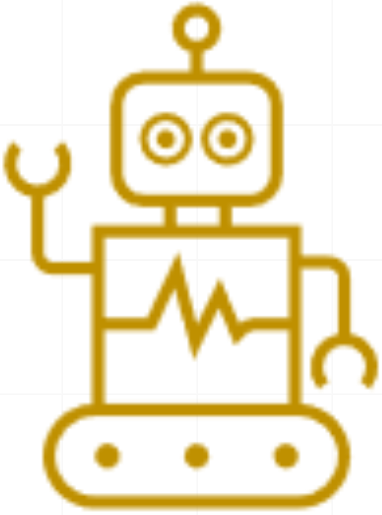}: Judge; \includegraphics[height=1em]{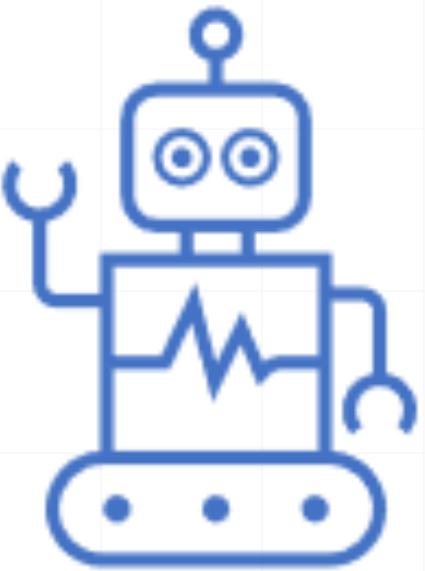}: Executor; \ding{192}: Initialize the initial condition list and the objective; \ding{193}: Explore new Conditions based on current condition list;
\ding{194}: Check if the new condition is correct; \ding{196}: Check if the objective can be achieved based on the current Conditions in the Condition list; \ding{197}: Designing steps for achieving the objective based on current Conditions; \ding{198}: Achieve the objective based on the designed steps.}
\label{fig:Overall_Structure}
\end{figure*}

The overall structure of MACM is shown in Figure~\ref{fig:Overall_Structure}.
We have designed an interactive system comprising three agents — Thinker, Judge, and Executor — to solve complex mathematical problems. \begin{itemize}
  \item \textbf{Thinker}: Responsible for generating new thoughts or ideas. This role involves creative thinking and the generation of novel solutions or approaches to problems.
  \item \textbf{Judge}: Tasked with evaluating the thoughts generated by the Thinker. The Judge assesses the viability and correctness of these ideas, ensuring that only the most logical and beneficial ones are pursued.
  \item \textbf{Executor}: Performs calculations or actions according to predefined steps. This role is focused on the practical application and implementation of the ideas approved by the Judge, turning abstract concepts into tangible outcomes.
\end{itemize}
 When a mathematical problem is input into our system, the Thinker initially sets up the \textbf{Condition List} and defines the final Objective based on the given problem.
After initialization, the Thinker mines new conditions conducive to the objective from the current Condition List, i.e., the Known Conditions. The Judge then assesses these newly mined conditions. If deemed correct, the Judge incorporates the new condition into the Condition List. Otherwise, the new condition is discarded.

Once all new Conditions have been reviewed, we obtain a revised Condition List. At this point, the Judge evaluates whether the current conditions are sufficient to achieve the objective. If the answer is False, the process reverts to step two for further mining of new conditions. In our experiments, we set a limit of five iterations; if the objective is not met after five rounds of mining, we consider the problem unsolvable. 
This prevents the program from entering an infinite loop. If the answer is True, the Thinker designs steps based on the Known Conditions in the current Condition List to achieve the Objective. Finally, the Executor performs calculations following these steps to produce the final result.

MACM achieves a high level of generalizability by abstracting conditions and objectives from each specific mathematical problem. Through a multi-agent interactive system, where the Thinker is responsible for ideation and design, the Judge for inspection and decision-making, and the Executor for computation, most potential errors in reasoning and calculation are eliminated. By repeatedly mining for conditions and adding the correct ones to the Condition List, MACM ensures depth in thinking, making it suitable for analyzing complex mathematical problems.

\subsection{Using Cases}
\begin{figure*}[h!]  
\centering 
\includegraphics[width=0.99\textwidth]{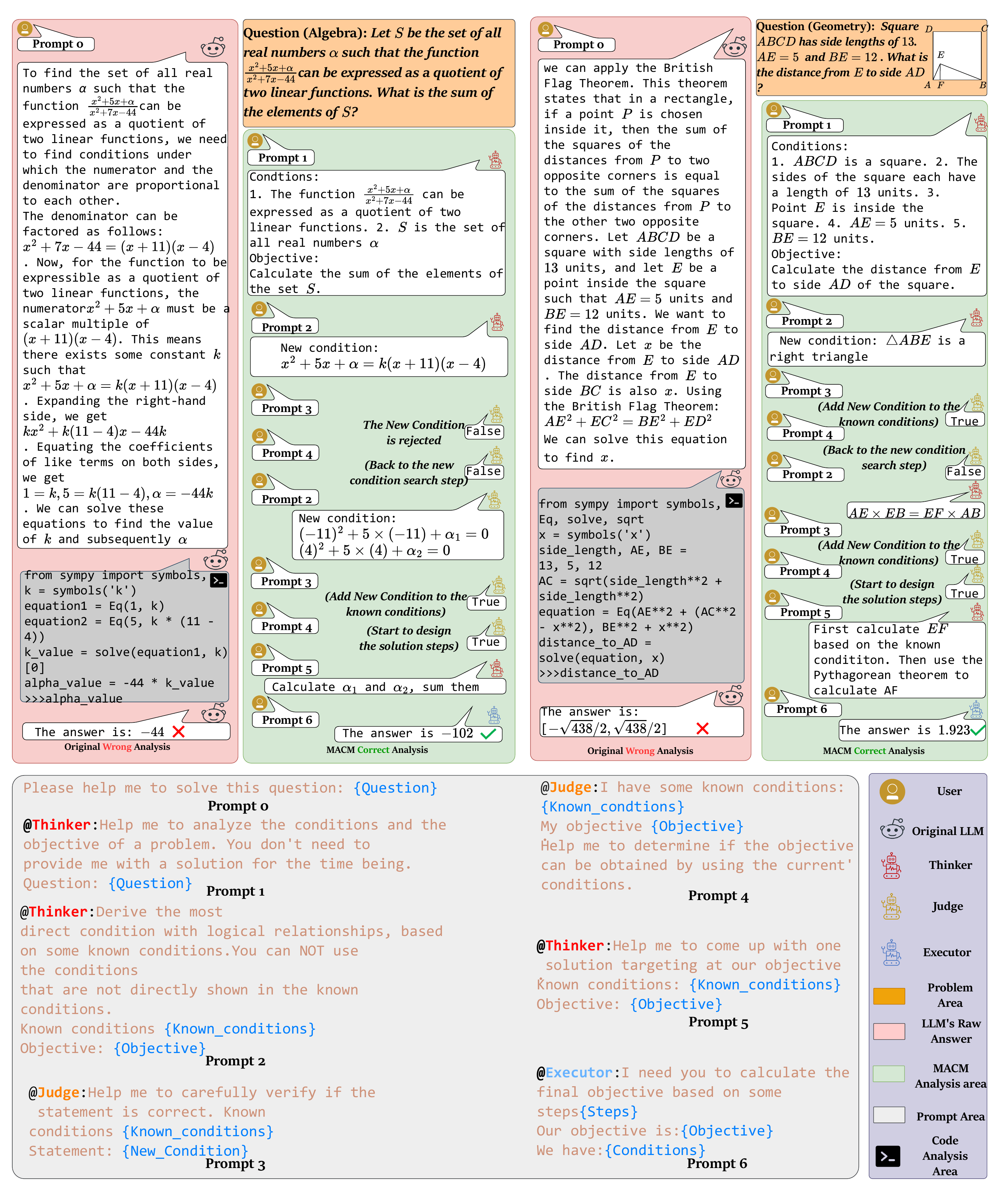} 
\caption{MACM's detailed analysis process for complex mathematical problems with specific prompts, illustrated with an \textbf{algebra problem (on the left)} and a \textbf{geometry problem (on the right)}. We use one set of prompts that can target different types of problems, with prompts 0-6 displayed in the \includegraphics[height=1em]{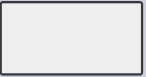} below the dialogue box. 
In these examples, MACM involves three steps: 1. Extracting conditions and the objective. 2. Iteratively identifying new conditions. 3. Solve the problem based on known conditions.}  
\label{fig:Usage_Cases}
\end{figure*}

Our prompts and use cases are shown in Figure \ref{fig:Usage_Cases}.
The figure demonstrates the specific process of MACM analyzing algebra and geometry problems. In these two examples, we have employed OpenAI's GPT-4 Turbo~\cite{openai_gpt4_turbo} as the intelligent agent. It is capable of performing calculations using code.
If the code runs incorrectly, it will continue to attempt modifications until the program can run or it reaches the maximum length. During the MACM analysis process of these two examples, GPT-4 Turbo is endowed with three roles: Thinker, Judge, and Executor.

In the first algebra problem:
\begin{framed}
Let \(S\) be the set of all real numbers \(\alpha\) such that the function \(\frac{x^2 + 5x + \alpha}{x^2 + 7x -44}\)can be expressed as a quotient of two linear functions. What is the sum of the elements of \(S\)?
\end{framed}
In GPT-4 Turbo's raw response, it arrived at an incorrect conclusion: \raisebox{0.0ex}{\fbox{\(x^2 + 5x + \alpha = k(x + 11)(x - 4)\)}}. This incorrect conclusion led to issues in the subsequent code design, ultimately resulting in an incorrect output.

In the MACM analysis process, the Thinker initially distinguishes conditions and objectives from the problem statement, then proceeds to uncover new conditions. Although the Thinker initially identifies the same incorrect condition as the original GPT-4 Turbo, the Judge successfully detects and rejects this error, preventing its addition to the Known Conditions. In the second round of condition discovery, the Thinker identifies two new conditions: \raisebox{0.0ex}{\fbox{\((-11)^2 + 5\times(-11) + \alpha_1 = 0\))}} and \raisebox{0.0ex}{\fbox{\((4)^2 + 5\times(4) + \alpha_2 = 0\)}} which the Judge verifies as correct and adds to the Known Conditions list. Subsequently, the Judge evaluates whether the Known Conditions are sufficient to achieve the objective. Receiving an affirmative response, the Thinker is then tasked with designing steps to reach the objectives based on the current Known Conditions. Finally, the Executor performs numerical calculations based on these steps to produce the result. 

In the second geometry problem:
\begin{framed}
Square \(ABCD\) has side lengths of 13 units. Point \(E\) lies in the interior of the square such that \(AE = 5\) units and \(BE = 12\) units. What is the distance from \(E\) to side \(AD\)?
\end{framed}
While the original response from GPT-4 Turbo provided the correct theoretical approach, it failed to clearly identify the relationships between various points in the problem. This led to incorrect relational expressions being listed, ultimately resulting in an incorrect final result.

During the MACM analysis process, the Thinker first clarifies the existing conditions and objectives of the geometry problem, then begins to uncover new conditions that are conducive to achieving the goal. Firstly, it discovers that: \raisebox{0.0ex}{\fbox{$\triangle ABE$ is a right triangle}}. This condition, after being verified by the Judge, is added to the current known conditions. The Judge then assesses whether the known conditions are sufficient to reach the objective. A 'False' return indicates that the current known conditions are insufficient, leading back to the new condition search step, where the Thinker continues to uncover other conditions. In the second condition discovery, the Thinker deduces another equation: \raisebox{0.0ex}{\fbox{$AE \times EB = EF \times AB$}}. This condition is deemed correct by the Judge and is added to the Known Condition List. The Judge then reassesses the sufficiency of the known conditions. Upon receiving an affirmative response, the Thinker starts planning the steps to solve the problem based on the current known conditions, and finally, the Executor performs the calculations to arrive at the result.

In the analysis of these two problems, MACM first extracts the specific conditions and the objective from their respective questions. This enables our method to directly use the obtained conditions and goals for prompt designing in subsequent decision-making processes, thereby enhancing the generalizability of our approach. Previous methods like ToT and GoT suffer from poorer generalizability due to the absence of such a setup. 

For example, in the 24-point game experiment with ToT, the lack of this setting necessitated the manual configuration of the following prompt in the subsequent evaluation phase:
\begin{framed}
Evaluate if given numbers can reach 24

(sure/likely/impossible)

\{input\}
\end{framed}
In MACM, the \raisebox{0.0ex}{\fbox{if given numbers can reach 24}} is obtained by the first step and the evaluation prompt is generalized to \raisebox{0.0ex}{\fbox{Evaluate \{objective\} \{input\}}}. Therefore, our method ensures a higher level of generalizability.

\subsection{Multi-Agents System}
It is crucial to clearly define the responsibilities and the format for answering questions of each agent. This ensures that each agent's role is distinct and their contributions are effectively integrated into the problem-solving process. The instructions provided for each agent are shown in the \raisebox{0.5ex}{\fbox{~~~~}}
 below.

\begin{framed}
\textbf{For Thinker}: \textit{You take the role of a thinker. I need you to help me gradually ponder over some problems following my instructions. You need to answer the question by using the following format: 
Based on Condition A and Condition B, we can get: C.}

\textbf{For Judge}: \textit{You take the role of a Judge. I need you to make judgments on some statements. You are only allowed to use the True or False as the final answer.}

\textbf{For Executor}: \textit{You take the role of a executor. I need you to calculate the final result based on the given conditions and steps.}
\end{framed}
If we do not differentiate agents and define a format for each agent to answer questions, it becomes challenging to standardize the responses generated by the original LLM. This lack of standardization can hinder the overall system's ability to function automatically. 

\begin{figure}[h]  
\centering 
\includegraphics[width=0.46\textwidth]{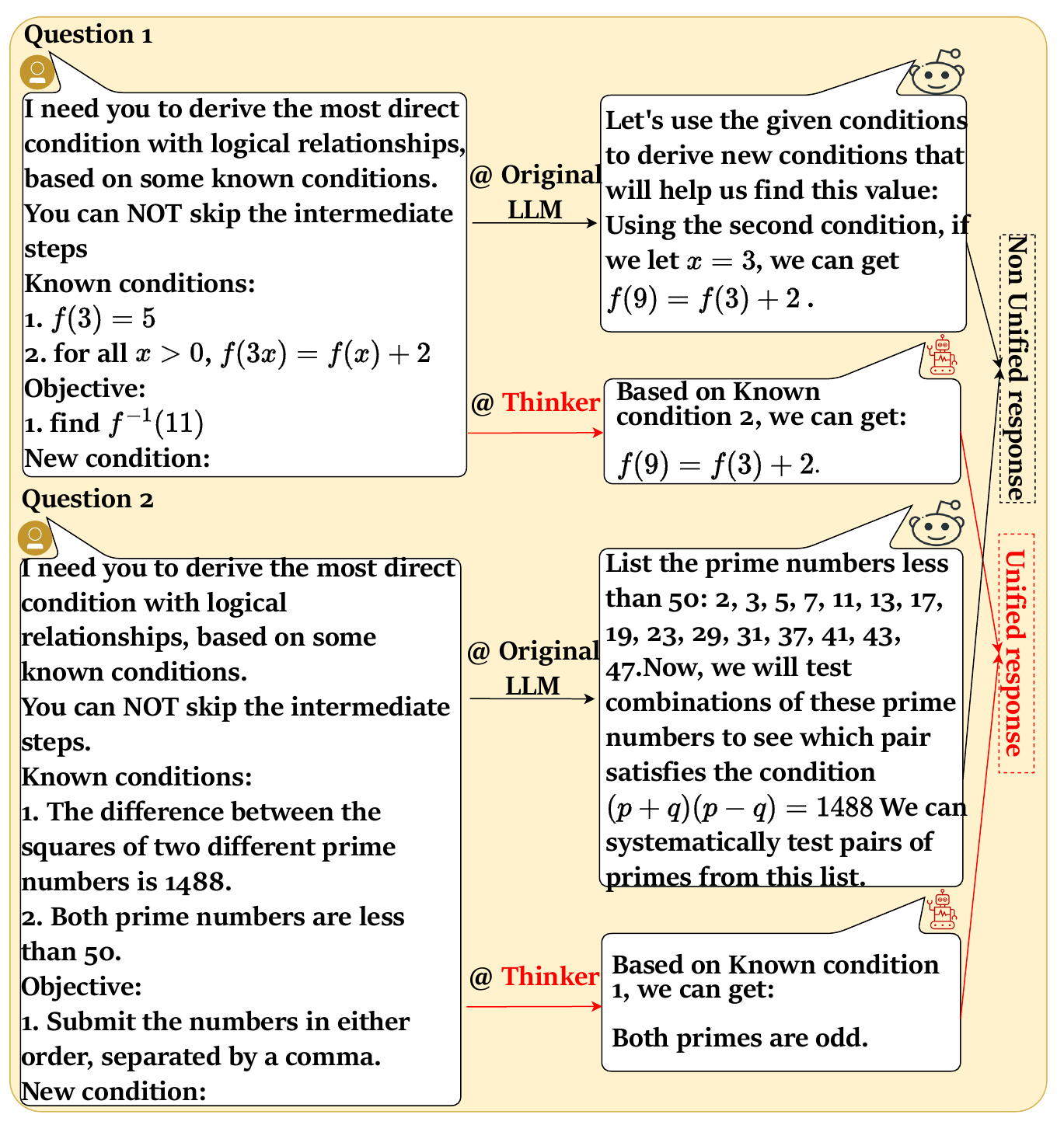} 
\caption{The differences in responses to various questions between an LLM with a defined identity (such as Thinker) and an LLM without a defined identity (like the original LLM). The Thinker consistently provides responses in the \textbf{same format}, while the original LLM produces responses in varying formats.\includegraphics[height=1em]{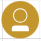}: User; \includegraphics[height=1em]{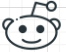}: Original LLM; \includegraphics[height=1em]{icon_image/Thinker.png}: Thinker.}  
\label{fig:Multi_Agents}
\end{figure}

An example illustrating this point is shown in the Figure~\ref{fig:Multi_Agents}.
For two different problems, GPT-4 Turbo may provide completely different response formats, making it challenging to process text and extract needed information. Previously, one solution was to use few-shot prompting for format uniformity. However, this method has two flaws: 1) the few-shot examples provided may influence the model's responses, and 2) it lacks generality, as numerous few-shot examples need to be independently designed for each problem. By clearly defining each agent's identity and giving specific formatting guidance, these issues can be avoided while maintaining consistency in response formats. As shown in the Figure~\ref{fig:Multi_Agents}, after presenting two questions to the Thinker, we received two responses in a uniform format: 
\begin{framed}
Based on known condition 2, we can get: $f(9) = f(3) + 2$
\end{framed}
 \begin{framed}
Based on known condition 1, we can get: Both primes are odd
\end{framed}
This approach greatly facilitates both the subsequent evaluation by the Judge and the extraction of new conditions.

\section{Experiment} 

We first evaluated the performance of MACM on the MATH dataset. To facilitate comparison with ToT and GoT, we then selected test cases where ToT and GoT were respectively applied—namely, the 24-point game and sequence sorting—and assessed MACM's performance in these two scenarios. Lastly, to gauge MACM's search efficiency, we examined the trade-off between accuracy and the total number of responses generated by GPT-4 turbo.
\label{sec:experiment}
\subsection{Performance on MATH benchmark}
The MATH dataset~\cite{hendrycksmath2021} includes a variety of mathematical problems. It offers seven types of mathematical problems, including geometry, algebra, probability theory, etc., with difficulty levels ranging from 1 to 5. We first tested the overall performance of MACM on the MATH dataset without distinguishing difficulty levels. Afterward, we specifically selected the most difficult mathematical problems from the MATH dataset for testing. The detailed experimental setup is presented in the Appendix~\ref{Appendix:Experimental_Setup}.
\begin{table*}[ht]
\caption{Accuracy(\%) comparison of GPT-4 Turbo on MATH dataset with different prompting Methods. $v$: voters} 
\label{tab:main_results} 
\centering
\resizebox{\textwidth}{!}{
\begin{tabular}{ccccccccc}
\hline
               & Algebra & \begin{tabular}[c]{@{}c@{}}Counting and \\ Probability\end{tabular} & Geometry & \begin{tabular}[c]{@{}c@{}}Intermediate \\ Algebra\end{tabular} & \begin{tabular}[c]{@{}c@{}}Number \\ Therogy\end{tabular} & Prealgebra & Precalculus & Overall \\ \hline
I-O            & 88.24   & 81.63                                                               & 45.11    & 66.67                                                           & 74.51                                                     & 81.82      & 71.15       & 72.78   \\
CoT            & 92.99   & 83.67                                                               & 42.02    & 68.07                                                           & 77.31                                                     & 82.07      & 74.18       & 74.36   \\
SC-CoT ($v = 5$) & 94.96   & 87.17                                                               & 50.14    & 71.99                                                           & 89.91                                                     & 86.75      & 79.67       & 80.12   \\
\cellcolor{lightgray}\textbf{MACM}  & \cellcolor{lightgray} 96.07        & \cellcolor{lightgray}97.95                                                               & \cellcolor{lightgray} 62.74        &\cellcolor{lightgray} 78.43                                                                &\cellcolor{lightgray}98.04                                                           &            \cellcolor{lightgray} 94.11&   \cellcolor{lightgray}88.46          &  \cellcolor{lightgray}   87.92    \\ \hline
\end{tabular}}
\end{table*}

In Table~\ref{tab:main_results}, we compared the accuracy of GPT-4 Turbo on the entire MATH dataset with various prompting methods. We found that compared to the original GPT-4 Turbo, MACM increased its accuracy by 20\%. Compared to CoT, the increase was 13.56\%, and compared to SC-CoT, it was 7.8\%. Among these, MACM led to the greatest improvement in accuracy for the original GPT-4 Turbo model on number theory problems, at 23.53\%. 
In geometry problems, although MACM has increased the accuracy of GPT-4 Turbo by 17.63\%, the final accuracy rate is still only 62.74\%. Upon analyzing the causes of errors, we found that many mistakes were due to GPT-4 Turbo's difficulty in accurately understanding the relationships between various geometric figures, thereby failing to design corresponding code to solve the problems. However, in algebra and number theory problems, MACM, by correcting the erroneous analysis of GPT-4 Turbo and helping it explore potential approaches, achieved accuracy rates of 96.07\% and 98.04\%, respectively. This fully demonstrates the effectiveness of MACM in solving mathematical problems.

\begin{figure}[h]  
\centering 
\includegraphics[width=0.46\textwidth]{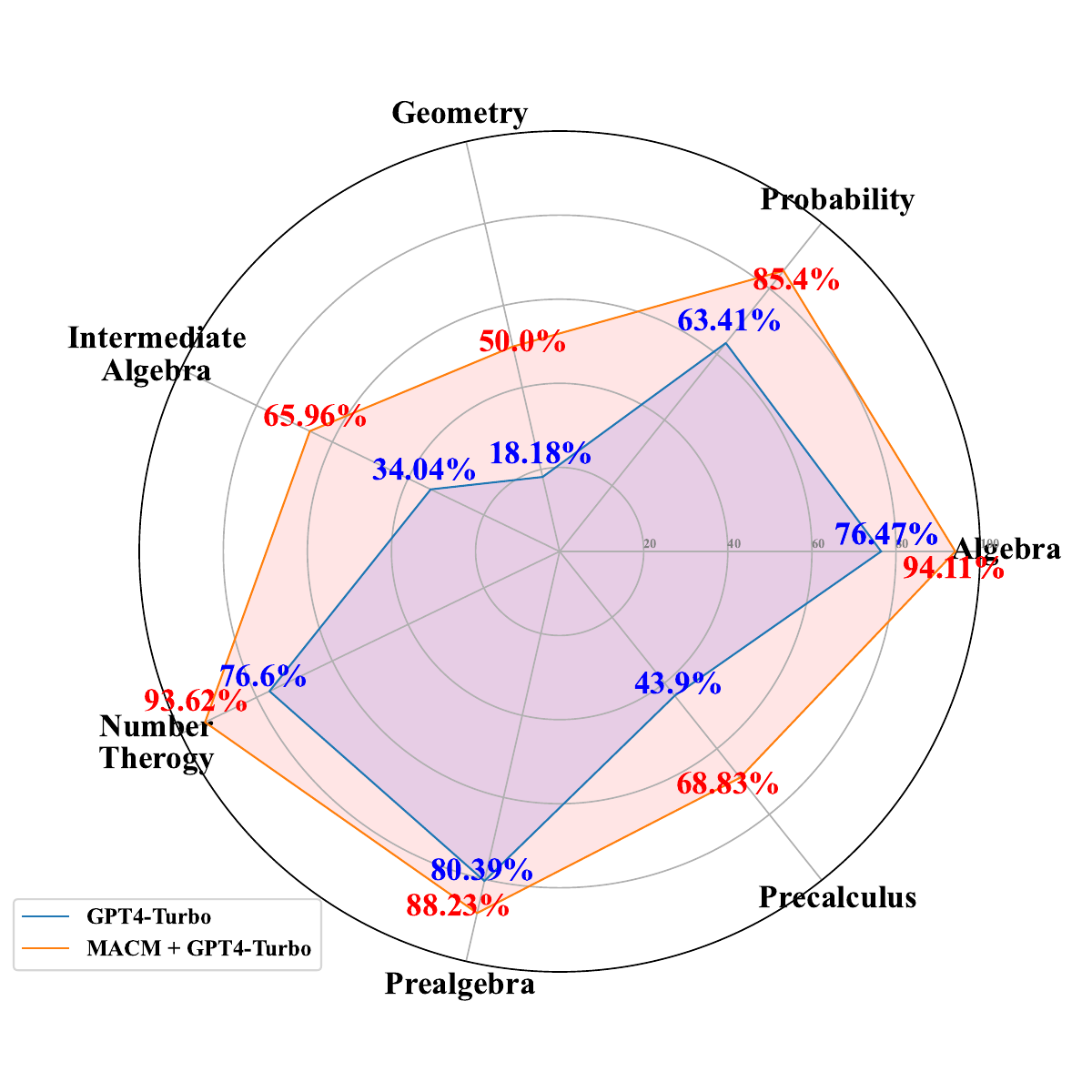} 
\caption{The performance comparison of GPT-Turbo with and without MACM on Level 5 problems of the MATH dataset.}  
\label{fig:Lv5_Comparision}
\end{figure}

In Figure~\ref{fig:Lv5_Comparision}, We focused on the ability of MACM to solve seven categories of Level 5 mathematics problems, including various knowledge areas such as geometry, algebra, probability theory, etc. As shown in the figure, MACM improved the accuracy of GPT-4 Turbo in all seven categories of level 5 problems. The two types of problems that saw the most significant improvement with MACM were the very categories where the original GPT-4 Turbo performed the worst: Geometry and Intermediate Algebra. The original GPT-4 Turbo had an accuracy rate of only 18.18\% on Geometry problems and 34.04\% on Intermediate Algebra problems. With the support of MACM, GPT-4 Turbo's accuracy rate in Geometry problems increased by 31.82\%, and in Intermediate Algebra problems, it increased by 31.92\%. This demonstrates MACM's effectiveness in solving difficult-level mathematical problems.

\subsection{Comparison with GoT and ToT}
Due to the lack of generalization of ToT and GoT prompting methods (See Appendix~\ref{Appendix:generalize} for the reason), we were unable to test them on the MATH benchmark. To compare MACM with them, we selected two mathematical problems where their methods are applicable: the 24-point game and sequence sorting. Among these, ToT tested the 24-point game, while GoT studied the sequence sorting problem. The detailed experimental setup is presented in the Appendix~\ref{Appendix:Experimental_Setup}.

\begin{table}[ht]
\caption{Accuracy(\%) comparison of different prompting Methods on 24-points game. $b$: Search breadth.} 
\label{tab:24_points_results} 
\centering
\begin{tabular}{cc}
\hline
Method                   & Accuracy (\%)\\ \hline
GPT-4 + IO~\cite{yao2023tree}               & 7.30   \\
GPT-4 + CoT~\cite{yao2023tree}              & 4.00    \\
GPT-4 + SC-CoT~\cite{yao2023tree} & 9.00   \\
GPT-4 + ToT ($b = 1$)~\cite{yao2023tree}      & 45      \\
GPT-4+ ToT ($b = 5$)~\cite{yao2023tree}       & 74      \\
\cellcolor{lightgray}GPT-3.5+ MACM            & \cellcolor{lightgray}67     \\
\cellcolor{lightgray}GPT-4 + MACM             & \cellcolor{lightgray}91     \\ 
\hline
\end{tabular}
\end{table}
In Table~\ref{tab:24_points_results}, We compared MACM with IO, CoT, SC-CoT, and ToT models on the 24-point game. When the model is GPT-4, MACM is 17\% higher than ToT ($b = 5$). Note that here, to ensure a fair comparison, we used the standard GPT-4 without any code capabilities, not GPT-4 Turbo. Additionally, with the support of MACM, GPT-3.5 also achieved an accuracy of 67\% in the 24-point game, which is higher than the GPT-4 model with ToT ($b = 1$) support. Upon analyzing the reasons for the improvement in accuracy, we found that MACM's Judge corrected many thoughts that were mistakenly evaluated in ToT, leading to GPT-4 choosing incorrect approaches. This correction process significantly contributed to the increase in accuracy.

\begin{table}[ht]
\caption{Accuracy(\%) comparison between GoT and MACM on sequence sorting problem (64 elements).}
\label{tab:Sorting} 
\centering
\begin{tabular}{cc}
\hline
Method                   & Accuracy (\%)\\ \hline
GPT-3.5 + GoT~\cite{besta2023graph}               & 89.06*   \\
\cellcolor{lightgray}GPT-3.5 + MACM             & \cellcolor{lightgray}92    \\
\hline
\end{tabular}
\end{table}
In Table~\ref{tab:Sorting}, We compare the ability of the GPT-3.5 model to sort 64 numbers under the support GoT and MACM. MACM is slightly higher than GoT by 2.94\%. * means that they present the results in a graph, which did not contain specific data. This result is estimated based on their graph.

\subsection{Trade-off Between Accuracy and LLM Queries}
In general, increasing the number of responses generated by large models theoretically leads to an improvement in accuracy. Each prompting method has parameters that can increase it, such as the length of the chain $l$ in CoT, the number of voters $v$ in SC-CoT, and the breadth $b$ and depth $d$ of the Tree in ToT. To measure the search efficiency of each method, we compared the relationship between the accuracy and the number of responses generated by GPT-4 Turbo on the MATH dataset. 

\begin{figure}[h]  
\centering 
\includegraphics[width=0.46\textwidth]{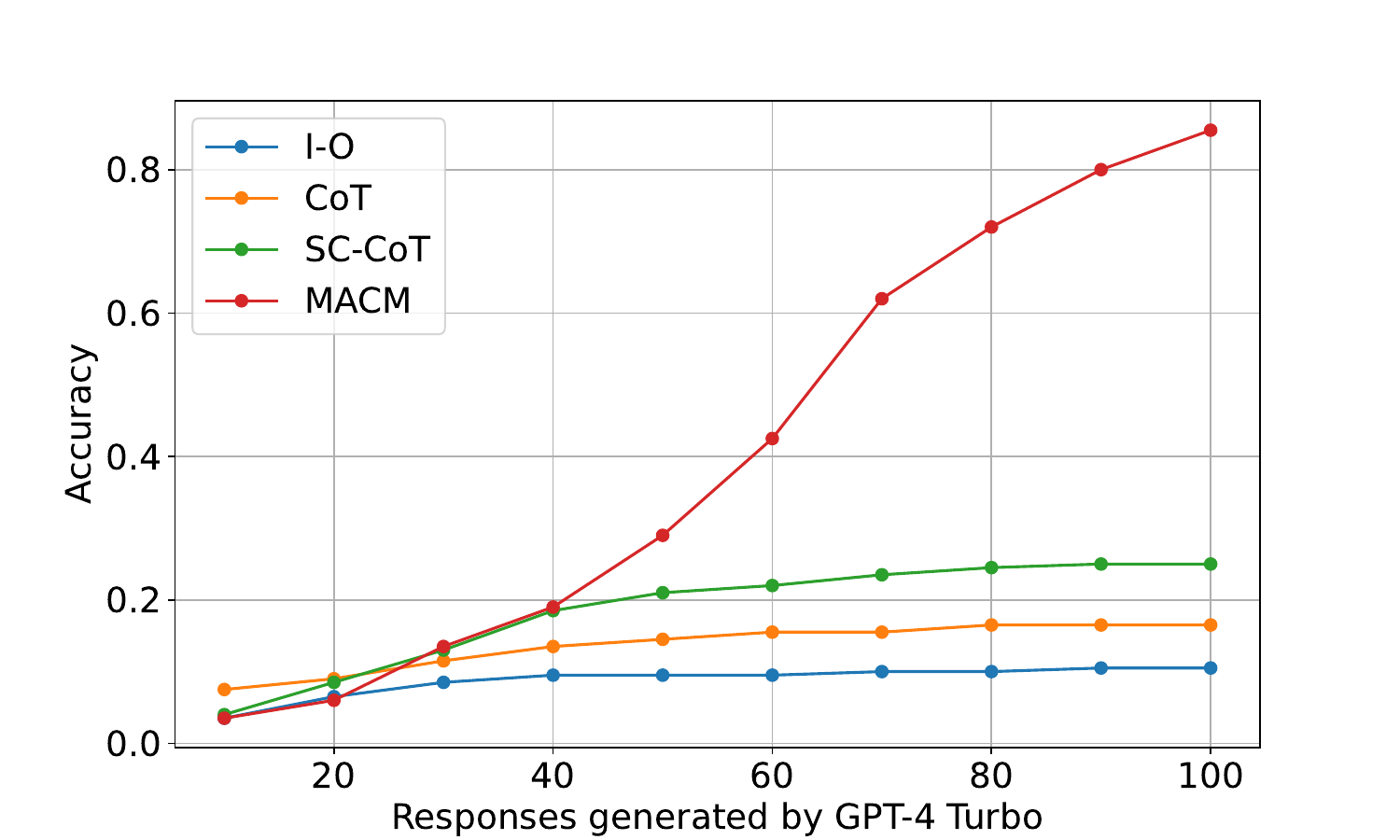} 
\caption{The trade-off between the accuracy and the responses generated by GPT-4 turbo.}  
\label{fig:Trade_Off}
\end{figure}

We first randomly selected 200 questions from the MATH dataset that the original GPT-4 Turbo model answered incorrectly. Then, we increased the number of answers generated by various prompting methods. For I-O prompting, we directly adjusted the model's response generation parameter $n$, which enables the large model to first generate $n$ responses and then select the best one from them. For the CoT method, we adjusted not only the parameter $n$ but also the length $l$ of the Chain. For the SC-CoT method, we built on the first two methods by adding an adjustment to the number of voters $v$. The result is shown in Figure~\ref{fig:Trade_Off}. 
The graph shows that although I-O, CoT, and SC-CoT only require simple queries to correct the original errors made by GPT-4 Turbo, their upper limits are not high. Even if we continue to increase the number of queries, they can only correct about 20\% of the original errors made by GPT-4 Turbo. In contrast, MACM can correct nearly 90\% of the original errors of GPT-4 Turbo when the number of queries is high. This is actually quite reasonable because the MACM structure is more complex, including multiple processes of mining conditions and checking. These processes allow the large model to gradually think and identify errors, thus significantly improving accuracy.

\section{Conclusion}
In this study, we introduce MACM, a new and generalizable prompting technique that significantly enhances the inferential capabilities of large language models on mathematical problems. Our set of prompts can be applied to different types of mathematical questions. Through comparisons in three experiments on the MATH dataset, the game of 24 points, and number sequence sorting, we have verified the superiority of our method over the original prompting methods. With the aid of our method, the accuracy of the GPT-4 Turbo model on the MATH dataset has increased by 15.14\%. Compared to SC-CoT, its accuracy has increased by 7.8\%. For the most challenging level 5 mathematical problems in the MATH dataset, its accuracy increased from 54.68\% to 76.73\%. In the game of 24 points, using the same GPT-4 model, MACM's accuracy is 17\% higher than that of ToT. At the same time, by comparing accuracy with the number of times the large model responds, we find that MACM has a higher limit; increasing the number of responses from the large model can significantly improve accuracy. These experiments demonstrate MACM's generalizability and its powerful error-correction capability for complex mathematical problems in original Large Language Models.

\section{Limitation and Discussion}

While MACM significantly enhances the accuracy of large language models in tackling complex mathematical challenges, it incurs the cost of multiple invocations of the large language model for inference, leading to increased problem-solving time. Additionally, our evaluations using the MATH dataset indicate limitations in effectively addressing geometry problems. Addressing these challenges necessitates further advancements in the model's own cognitive capabilities. A proposed strategy involves employing prompting methods like MACM to assist the LLM in eliminating incorrect responses. This approach enables the creation of expansive, high-quality datasets, which are otherwise challenging to compile manually, and subsequently refining the LLM with these datasets. Through this iterative process, the model's intrinsic intelligence is progressively augmented. 
This research direction will constitute our future work.

\section*{Impact Statements}
This paper presents work whose goal is to advance the field of Machine Learning. There are many potential societal consequences of our work, none which we feel must be specifically highlighted here.
\bibliography{bio}
\bibliographystyle{icml2024}

\newpage
\appendix
\onecolumn
\section{
Why is the generalizability of ToT and GoT limited}
\label{Appendix:generalize}
This section demonstrates specific examples of using ToT and GoT to further illustrate why their generalizability is limited.

ToT conducted three sets of experiments in the original study, requiring specially designed prompts for each. The official implementation on their GitHub page https://github.com/princeton-nlp/tree-of-thought-llm/tree/master/src/tot/prompts includes the specific prompts set up for each experiment. Taking the 24-point game as an example, specific prompts such as propose prompt, value prompt, and value last step prompt were required (lines 51 to 134 in game24.py). During ToT's operation, the LLM executes traversal searches, voting, filtering, etc., based on these written prompts. The authors also mention in the ToT readme section \textit{How to Add a New Task }(https://github.com/princeton-nlp/tree-of-thought-llm?tab=readme-ov-file\#how-to-add-a-new-task) that setting up task-specific prompts is necessary for different problems, further illustrating the limited generalizability of ToT and GoT due to the need for task-specific prompt engineering.

GoT faces the same issue, with their original paper conducting experiments in four tasks: Sorting, Set Operations, Keyword Counting, and Document Merging. For each type of problem, specific prompts must be set up on their official GitHub. Taking Sorting as an example, the specific prompts for sorting are displayed in https://github.com/spcl/graph-of-thoughts/blob/main/examples/sorting/example\_prompts\_sorting\_032.md. They provide the LLM with the instruction: \textit{Split the following list of 32 numbers into 2 lists of 16 numbers each, the first list should contain the first 16 numbers and the second list the second 16 numbers. Only output the final 2 lists in the following format without any additional text or thoughts!} This instruction is clearly tailored to this specific problem, illustrating the limited generalizability of GoT due to the necessity for problem-specific prompt engineering, similar to ToT.

The requirement to tailor prompts for each specific problem limits the generalizability of ToT and GoT to broader issues. MACM successfully addresses this challenge.

\section{Experimental Setup}
\label{Appendix:Experimental_Setup}
\textbf{For the experiments on the MATH dataset:} We utilized the GPT-4 Turbo model (between January 1, 2024, and February 1, 2024) to test MACM's performance on the MATH dataset. For tests that did not distinguish by difficulty, we randomly selected one-third of the questions from the MATH dataset for evaluation. For the high-difficulty tests, we extracted all questions with a difficulty level of 5 and randomly selected half of the questions from each category for testing. The experiment are performed by using I-O, CoT, SC-CoT, and MACM methodologies. For all prompting methods, we standardized the number of responses $n$ generated by GPT-4 Turbo to $1$, $Top_k = 1$ , and the temperature $t = 0.7$. For CoT, we set the maximum length of the chain $l = 5$, for SC-CoT, the number of voters $v = 5$, and the maximum length of the chain $l = 5$. For these three methods, we consistently maintained max\_tokens at $512$. For MACM, we kept the thinker's max\_tokens at $512$, the judge's max\_tokens at $4$, and the executor's max\_tokens at $256$.

\textbf{For the 24-point game experiment}: We sourced data from 4nums.com, which offers 1,362 games ranked from easy to hard based on the time it takes humans to solve them. We focused on a subset of these games, specifically those ranked 901 to 1,000 (The same as ToT), to test on relatively difficult challenges. Success for each task is defined as producing a valid equation that results in 24, utilizing each of the input numbers exactly once. The performance metric is the success rate across these 100 challenging games.We utilized the GPT-4 and GPT-3.5 model (between January 1, 2024, and February 1, 2024) to perform the experiments. The MACM configuration for this experiment includes setting the number of responses generated by the model $n = 1$, $Top_k = 1$, temperature $t = 0.7$, with the thinker's max\_tokens at $512$, the judge's max\_tokens at $4$, and the executor's max\_tokens at $256$.

\textbf{For the sequence sorting experiment}: We randomly generated 100 sequences, each containing 64 elements, for testing. We utilized the GPT-3.5 model (between January 1, 2024, and February 1, 2024) to perform the experiments. The MACM configuration for this experiment includes setting the number of responses generated by the model $n = 1$, $Top_k = 1$, temperature $t = 0.7$, with the thinker's max\_tokens at $512$, the judge's max\_tokens at $4$, and the executor's max\_tokens at $256$.
\end{document}